\documentclass{article}

\usepackage{arxiv}

\usepackage[utf8]{inputenc} 
\usepackage[T1]{fontenc}    
\usepackage{hyperref}       
\usepackage{url}            
\usepackage{booktabs}       
\usepackage{amsfonts}       
\usepackage{nicefrac}       
\usepackage{microtype}      
\usepackage{lipsum}
\usepackage{amsmath}
\usepackage[ruled]{algorithm2e}

\title{Randomized Ablation Feature Importance}

\author{
 Luke Merrick \\
  Fiddler Labs \\
  \texttt{luke@fiddler.ai} \\
}

\DeclareMathOperator*{\argmin}{arg\,min}

\newcommand{\Reals}{\ensuremath\mathbb{R}}
\newcommand{\Ex}[2][]{\ensuremath{\mathop\mathbb{E}_{#1}\left[ {#2} \right]}}
\newcommand{\Prob}[2][]{\ensuremath{\mathop\mathsf{Pr}_{#1}\left[ {#2} \right]}}

\newcommand{\fimport}{\ensuremath \Delta R}
\newcommand{\hatfimport}{\ensuremath \Delta \hat{R}}
\newcommand{\deltaloss}{\ensuremath \Delta \ell}
\DeclareMathOperator{\randsample}{random\_sample}
\DeclareMathOperator{\marg}{marginal}
\DeclareMathOperator{\Sem}{SEM}
\pdfoutput=1

\begin{document}
\maketitle

\begin{abstract}
Given a model $f$ that predicts a target $y$ from a vector of input features $\pmb{x} = x_1, x_2, \ldots, x_M$, we seek to measure the importance of each feature with respect to the model's ability to make a good prediction. To this end, we consider how (on average) some measure of goodness or badness of prediction (which we term ``loss'' $\ell$), changes when we hide or \emph{ablate} each feature from the model. To ablate a feature, we replace its value with another possible value randomly. By averaging over many points and many possible replacements, we measure the importance of a feature on the model's ability to make good predictions. Furthermore, we present statistical measures of uncertainty that quantify how confident we are that the feature importance we measure from our finite dataset and finite number of ablations is close to the theoretical true importance value.
\end{abstract}


\section{Introduction}
The idea we call ``randomized ablation feature importance'' is not new. Indeed, its inception dates back to at least 2001, when a variant of this technique was introduced as the ``noising'' of variables to better understand how random forest models use them \cite{breiman2001random}. Recently, however, this technique has seen a resurgence in usage and variation. For example, the popular Scikit-learn package \cite{scikit-learn} has, at the time of this writing, included an implementation of this technique in their development build. Titled ``permutation importance,'' this feature will likely ship in their next release. On the theoretical side, a recent work proposing a framework of ``model class reliance'' has termed a variant of the randomized ablation feature importance ``model reliance'' and used it as the backbone of their technique \cite{fisher2018all}.

Although some works like \cite{strobl2008conditional} offer a rigorous statistical treatment of the technique, we have not been able to find a theoretically precise explanation of this technique that is also concise and clear. Through this work, we seek to fill the gap. We develop an intuitive and clear theoretical formulation of this technique as it relates to the classic statistical learning problem statement. We find that the notion of randomized ablation feature importance actually fits in quite nicely with the mathematics of risk minimization in supervised learning, a connection we find helpful to our intuition. Additionally, through this formulation, we identify an opportunity to estimate the uncertainty inherent in this randomized algorithm's outputs. To the best of our knowledge, current formulations and implementations of this technique do not include these confidence measurements, making this perhaps a novel and useful contribution in its own right.

\section{Preliminaries: the mathematics of supervised learning}
We find it instructive to establish a mathematically formal statement of randomized ablation feature importance, such that the measurement used in practice can be considered an estimation of this quantity. In order to define this theoretical importance measure, we need to introduce some mathematical preliminaries. We adopt the standard mathematical framework of \emph{supervised learning}, following the typical assumption that some \emph{fixed but unknown} multivariate distribution $\mathcal{D}$ randomly generates our data. Our goal is to predict a target variable $Y$ based upon the multivariate predictive variable $\pmb{X} = X_1, X_2, \ldots, X_M$, where $(\pmb{X}, Y) \sim D$. We use a learning algorithm to output a model function $f$, which represents our best guess of $Y$ based upon $\pmb{X}$. Typically our goal is that this learning algorithm will, using a finite sample of $(\pmb{x}, y)$ pairs, identify a model $f$ such that $f(\pmb{X}) \approx \Ex{Y | \pmb{X}}$.\footnote{Potentially with some smoothing if $\Ex{Y | \pmb{X}}$ is very jumpy. See \cite{gao2017interpretability} for a discussion of the impossibility of exact calibration between the model and the conditional probability.}

\subsection{Risk}
To understand how good our model is, we choose some loss function $\ell: \Reals \times \Reals \rightarrow \Reals$. We then evaluate our model of by measuring the expected loss (i.e. the ``average badness''), which we call the "risk" of the model: 

\begin{equation}
    R(f) = \Ex{\ell(f(\pmb{X}), Y)}
\end{equation}

\subsection{Unknown distribution, finite data}\label{sec:finite_data}
Although we imagine learning in terms of random variables, in practice we don't actually know the true nature of our data distribution $\mathcal{D}$. Instead, we use a concrete dataset consisting of $N$ samples of $(\pmb{X}, Y) \sim \mathcal{D}$:
\begin{equation*}
    S = \left( (\pmb{x}^{(j)}, y^{(j)}) \right)_{j=1}^N
\end{equation*}

Since we do not know the nature of $\mathcal{D}$, we also don't know $R$, but we can approximate it using our dataset. Doing so gives us the ``empirical'' or (as we shall call it in this paper) \emph{fixed data} risk:
\begin{equation}\label{eq:fixed_data_R}
    R^{FD}(f) = \frac{1}{N} \sum_{j=1}^{N} \left[ \ell \left( f \left( \pmb{x}^{(j)} \right), y^{(j)} \right) \right]
\end{equation}

\subsection{Sidenote: risk minimization}\label{sec:risk_minimization}
Risk not only plays a role in evaluating models, but also in learning them. The risk minimization approach to supervised learning calls for selecting the model that represents our best guess of the risk minimizer $f^* = \argmin_f~R(f)$. 
In this paper we focus on using risk to evaluate a model, but much of the mathematics is borrowed from the mathematics of risk minimization. Those familiar with the theory of risk minimization may find it a useful conceptual anchor for this discussion of randomized ablation feature importance.

\section{Defining randomized ablation feature importance}
For a specific model, we want to know how important each dimension of the input is. In other words, how much does the feature $X_i$ matter to the model $f$? One way to measure this is to observe how much the risk increases when we ``mess up'' the $i$th input. This ``messing up'' can be accomplished through \emph{randomized ablation}, in which we replace $X_i$ with a random variable $Z_i$ that is distributed according to the marginal distribution of $X_i$. Since $Z_i$ is distributed according to the marginal distribution of $X_i$, it is statistically independent of $Y$ (i.e. $\Prob{Y | Z_i} = \Prob{Y}$)) and thus useless for predicting $Y$.

We define the randomly ablated risk for the $i$th feature as follows:
\begin{equation}
    R_{\setminus i}(f) = \Ex{\ell(f(X_1, X_2, ..., Z_i, ..., X_M), Y)}
\end{equation}

The change in risk caused by ablation serves as a measurement of how important the ablated feature is to the model's ability to make good predictions. In other words, it gives us the \emph{randomized ablation feature importance}:
\begin{equation}\label{eq:delta}
    \fimport_i = R_{\setminus i}(f) - R(f) = \Ex{\ell(f(X_1, X_2, ..., Z_i, ..., X_M), Y) - \ell(f(\pmb{X}), Y)}
\end{equation}

\subsection{The ``fixed-data'' alternate formulation}\label{sec:alternate}
In practice we often focus not on the idea of true risk, but rather on the fixed data approximation of risk we actually observe. Regardless of whether we define the true randomized ablation feature importance in terms of the true risk or the fixed data risk, our estimate will be the same in practice (we demonstrate this in Section~\ref{sec:estimation}). However, this equivalence does not hold for our notion and measure of our estimate's uncertainty. Rather, our treatment of uncertainty must depend on the formulation of \emph{true} feature importance we consider. Accordingly, we now introduce the alternate ``fixed-data'' definitions of randomly ablated risk and randomized ablation feature importance:
\begin{equation}
    \label{eq:fixed_data_R_ablated}
        R^{FD}_{\setminus i}(f) = \Ex{ \frac{1}{N} \sum_{j=1}^{N} \left[ \ell \left( f \left( x_1^{(j)}, x_2^{(j)}, \ldots, Z_i, \ldots, x_M^{(j)} \right), y^{(j)} \right) \right]}
\end{equation}

\begin{equation}\label{eq:fixed_data_delta}
    \begin{aligned}[b]
        \fimport^{FD}_i &= R^{FD}_{\setminus i}(f) - R^{FD}(f)  \\
        &= \Ex{ \frac{1}{N} \sum_{j=1}^{N} \left[ \ell \left( f \left( x_1^{(j)}, x_2^{(j)}, \ldots, Z_i, \ldots, x_M^{(j)} \right), y^{(j)} \right) \right] } - \frac{1}{N} \sum_{j=1}^{N} \left[ \ell \left( f \left( \pmb{x}^{(j)} \right), y^{(j)} \right) \right] \\
        &= \Ex{\frac{1}{N} \sum_{j=1}^{N} \left[ \ell \left( f \left( x_1^{(j)}, x_2^{(j)}, \ldots, Z_i, \ldots, x_M^{(j)} \right), y^{(j)} \right) - \ell \left( f \left( \pmb{x}^{(j)} \right), y^{(j)} \right) \right]}
    \end{aligned}
\end{equation}
Even though both of these quantities rely on our fixed dataset, they still involve an expectation over $Z_i$, and thus they still represent theoretical quantities that we can only estimate in practice.

\subsection{Simplifying notation}
For the sake of concise notation, let's define the \emph{ablation loss delta}:
\begin{equation*}
    \deltaloss (f, \pmb{x}, y, z_i) = \ell(f(x_1, x_2, \ldots, z_i, \ldots, x_M), y) - \ell(f(\pmb{x}), y)
\end{equation*}

We can now rewrite $\fimport_i$ and $\fimport^{FD}_i$ as follows:

\begin{equation}\label{eq:concise_delta_R}
   \fimport_i = \Ex{\deltaloss(f, \pmb{X}, Y, Z_i)}
\end{equation}

\begin{equation}\label{eq:concise_delta_R_FD}
   \fimport^{FD}_i = \Ex{ \frac{1}{N} \sum_{j=1}^{N} \left[\deltaloss(f, \pmb{x}^{(j)}, y^{(j)}, Z_i) \right]}
\end{equation}

\section{Estimating randomized ablation feature importance}\label{sec:estimation}
In practice, we don't know the true distribution $\mathcal{D}$, but we have access to our dataset $S$, a finite sample of $(\pmb{x}, y)$ pairs. We use this to empirically approximate the randomized ablation feature importance. To handle ablations of the $i$th feature, we approximate $Z_i \sim \marg(X_i)$ by randomly sampling $N \times K$ values from our dataset:
\begin{equation}
    T =\left( z_i^{(k)} \right)_{k=1}^{N\times K} = \randsample \left( \left( x_i^{(j)} \right)_{j=1}^N~,~N \times K \right)
\end{equation}

The \emph{empirical randomized ablation feature importance}, is then computed as follows:

\begin{equation}\label{eq:empirical_delta}
    \begin{aligned}[b]
        \hatfimport_i &= \frac{1}{N \times K} \sum_{k=1}^{K} \sum_{j=1}^{N}  \ell(f(\pmb{x}^{(j)}), y^{(j)}) - \ell(f(x_1^{(j)}, x_2^{(j)}, \ldots, z_i^{(k \times j)}, \ldots, x_M^{(j)}), y^{(j)}) \\
        &= \frac{1}{N \times K} \sum_{k=1}^{K} \sum_{j=1}^{N}  \Delta \ell(f, \pmb{x}^{(j)}, y^{(j)}, z_i^{(k \times j)})
    \end{aligned}
\end{equation}

Under a few assumptions, this empirical randomized ablation feature importance is an estimate of the true feature importance as defined in both the random-variable and fixed-data formulations. To see this, let us first consider the fixed-data formulation. A sampling approximation of Equation~\ref{eq:concise_delta_R_FD} is achieved by averaging the expression within the expected value as it is evaluated on independent samples of $Z_i$. In constructing $T$, we implicitly assume that sampling from the empirical distribution of $Z_i$ given by our dataset is a good approximation of sampling from $Z_i$. Under this assumption, it follows that Equation~\ref{eq:empirical_delta} is an average of $K$ independent samples of the expression within Equation~\ref{eq:concise_delta_R_FD}, and thus a sampling approximation of this quantity.


For random-variable formulation of randomized ablation feature importance, we need a sampling approximation for Equation~\ref{eq:concise_delta_R}. It is easiest to begin by considering the case where $K = 1$. Since our dataset is a sequence of independent samples of $(\pmb{X}, Y) \sim \mathcal{D}$, if we again assume that $T$ is a good approximation of a sequence of independent samples of $Z_i$, then Equation~\ref{eq:empirical_delta} is an average of random samples of the expression within Equation~\ref{eq:concise_delta_R}, and thus a sampling approximation of $\fimport$. For $K > 1$, there is an additional implicit assumption. We assume that sampling from the cross product of the empirical distribution of $(\pmb{X}, Y)$ and the empirical distribution of $Z_i$ is a good approximation of sampling from the cross products of the true distributions of these variables. Equation~\ref{eq:empirical_delta} is a sampling approximation of Equation~\ref{eq:concise_delta_R} for all values of $K$ under which we are willing to make this additional assumption.

\section{Confidence intervals}\label{sec:cis}
In both the random-variable and fixed-data formulations, empirical randomized ablation feature importance is an estimate of the mean of a random variable from a finite sample. As a result, we can construct confidence intervals (CIs) for this quantity \cite{tabogalectures}. The random-variable formulation confidence interval will capture the confidence of our estimate over the theorized data distribution, while the fixed-data formulation CI will account for the uncertainty of the permutation under the assumption that the data sample is fixed.


Using the maximum likelihood estimate of variance, we first estimate the sample Standard Error of the Mean (SEM) as follows:
\begin{equation}\label{eq:sem}
    \Sem \left( \left(\deltaloss^{(s)} \right)_{s=1}^N \right) = \sqrt{\frac{\hat{\sigma}^2}{N}}
\end{equation}
Where:
\begin{align*}
    \hatfimport &= \frac{1}{N} \sum_{s=1}^{N} \deltaloss^{(s)} \\
    \hat{\sigma}^2 &= \frac{1}{N} \sum_{s=1}^{N} \left( \deltaloss^{(s)} - \hatfimport \right)^2
\end{align*}

For the random-variable formulation CI, each ablation loss delta $\deltaloss^{(s)}$ corresponds to a sampled ablation loss delta $\Delta \ell(f, \pmb{x}^{(j)}, y^{(j)}, z_i^{(k \times j)})$, with $s = k\times j$. For the fixed-data formulation each $\deltaloss^{(s)}$ corresponds to an average $\frac{1}{N} \sum_{j=1}^{N}  \Delta \ell(f, \pmb{x}^{(j)}, y^{(j)}, z_i^{(k \times j)})$, where $s = k$.

To use the SEM to construct confidence intervals around our feature importance estimates, we multiply the sample SEM by the number of standard deviations that contain the desired total probability density in a $t$-distribution. For a simple example, let's approximate a $t$-distribution with a normal distribution. We multiply the SEM by 1.96 for a 95\% confidence interval (an excellent approximation for sample sizes of 100 or more), since the total probability mass for a normal distribution is 0.95 within 1.96 standard deviations of the mean. This gives us the following 95\% confidence interval on feature importance:

\begin{equation}
    \left[\hatfimport - 1.96 \times \sqrt{\frac{\hat{\sigma}^2}{N}}, \hatfimport + 1.96 \times \sqrt{\frac{\hat{\sigma}^2}{N}} \right]
\end{equation}

\section{Conclusion}
We hope that this treatment of the theory and practice of randomized ablation feature importance serves as a helpful guide to those who seek to use this technique in their own work. By establishing a connection to the concept of \emph{risk} from statistical learning theory, we have added a measure of theoretical rigor to an otherwise ``rule-of-thumb'' method. At the same time, we have provided a methodology for the assessment of uncertainty that we hope will benefit those who apply this technique in practice. Above all, however, we hope that this provides a helpful and thorough introduction into this two-decade-old technique.

\bibliographystyle{unsrt}  
\bibliography{references}  

\end{document}